\pdfoutput=1

\documentclass[11pt]{article}

\usepackage{acl}

\usepackage{times}
\usepackage{latexsym}

\usepackage[T1]{fontenc}

\usepackage[utf8]{inputenc}

\usepackage{microtype}
\usepackage{amsmath}
\usepackage{graphicx}

\title{Infusing Knowledge from Wikipedia to Enhance Stance Detection}

\author{
  Zihao He$^{1,2}$ \quad Negar Mokhberian$^{1,2}$ \quad Kristina Lerman$^{1}$ \\
  $^1$Information Sciences Institute, University of Southern California\\
  $^2$Department of Computer Science, University of Southern California \\
  \texttt{ \{zihaoh,nmokhber\}@usc.edu}   \quad  \texttt{lerman@isi.edu}\\
  }

\begin{document}
\maketitle
\begin{abstract}
Stance detection infers a text author's attitude towards a target. 
This is challenging when the model lacks background knowledge about the target.
Here, we show how background knowledge from Wikipedia can help enhance the performance on stance detection. We introduce \textbf{W}ikipedia \textbf{S}tance Detection BERT (WS-BERT) that infuses the knowledge into stance encoding. 
Extensive results on three benchmark datasets covering social media discussions and online debates indicate that our model significantly outperforms the state-of-the-art methods on target-specific stance detection, cross-target stance detection, and zero/few-shot stance detection.\footnote{Code and data are publicly available at \url{https://github.com/zihaohe123/wiki-enhanced-stance-detection}.}

\end{abstract}

\section{Introduction}
\begin{figure}[ht]
    \centering
    \includegraphics[width=0.38\textwidth]{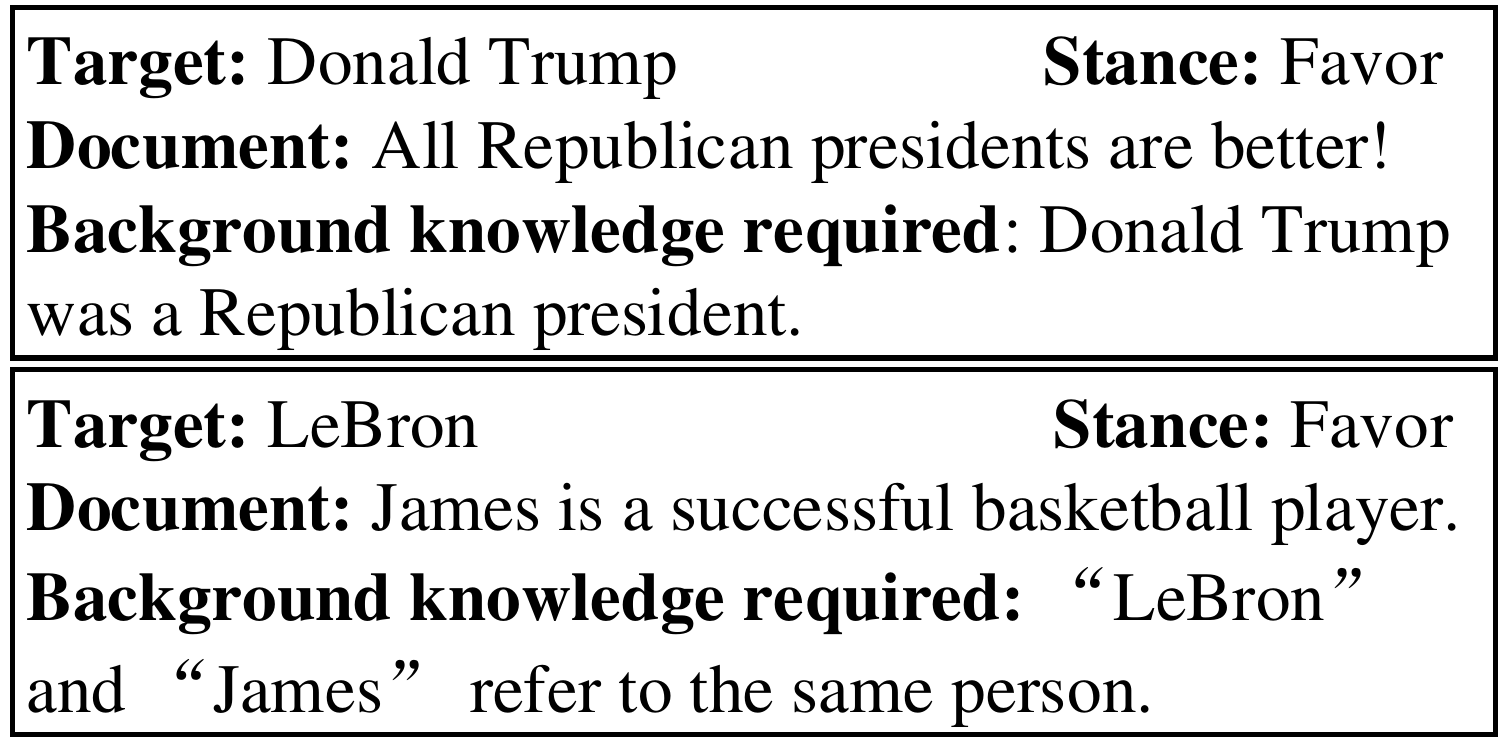}
    \caption{Two examples of stance detection where background knowledge is required.}
    \label{fig:examples}
\end{figure}

Stance detection aims to automatically identify author's attitude or standpoint (favor, neutral, against) towards a specific target or topic using text as evidence~\cite{mohammad2016semeval, augenstein2016stance, jang2018explaining, somasundaran2010recognizing, stefanov2020predicting}. 
To precisely capture the stance towards a target, background knowledge about the target is often necessary, especially in cases where the text does not explicitly mention the target, as shown in Figure \ref{fig:examples}.
People have wide-ranging background knowledge regarding various targets and use it to infer the implicit stance in a statement. However, machines by default do not have such knowledge and previous works on stance detection \cite{allaway2020zero, allaway2021adversarial, liang2021target, augenstein2016stance, siddiqua2019tweet, sun2018stance, li2021improving, hardalov2021cross} fail to incorporate such knowledge in modeling stances.

In this paper, we propose to utilize background knowledge from Wikipedia about the target as a bridge to enable the model's deeper understanding of the target, thus improving its performance on stance detection. 
We crawl the Wikipedia pages for the targets and use them as external textual information. To infuse this information into stance detection, we propose \textbf{W}ikipedia \textbf{S}tance Detection BERT (WS-BERT), which integrates the representation of Wikipedia knowledge into that of documents and targets. Depending on the textual style of the documents, we introduce two variants of WS-BERT.
We conduct a comprehensive set of experiments on three recently published benchmark datasets for stance detection that include social media discussions and online debates, covering three sub-tasks of stance detection: target-specific stance detection, cross-target stance detection, and zero/few-shot stance detection. Significant improvements over the state-of-the-art methods on all datasets and sub-tasks demonstrate the superiority of our model in terms of effectiveness and broad applicability.

\textbf{Related Work.} 
\citet{baly2018predicting, baly2020written} use Wikipedia pages of a news medium as an additional source of information to predict the factuality and bias of the medium. However, they use static pretrained BERT~\cite{devlin2018bert}  embeddings of the Wikipedia pages without finetuning, failing to align the pretrained embeddings to the domain of the target task.
\citet{hanawa2019stance} first propose to make use of the external knowledge from Wikipedia for stance detection; however, the authors only consider the promote/suppress relations between the texts and Wikipedia, which require a large amount of manual annotations to extract; in addition, a substantial amount of knowledge that is not captured by such relations is ignored; in contrast, WS-BERT utilizes the original Wikipedia textual knowledge and does not proactively exclude any information.
\citet{zhang2020enhancing} propose SEKT to extract external word-level semantic and emotion knowledge, which fails to capture the global relationship between the document and the target; moreover, such a model is designed for cross-target stance detection and is hardly applicable to target-specific and zero/few-shot stance detection. 
\citet{liu2021enhancing} utilize commonsense knowledge from a knowledge graph by extracting the two-hop paths between entities in the targets and in the documents; however, the existence of such paths do not always hold true and we found that a well-finetuned BERT without external knowledge can achieve performance comparable with it, as shown in Section \ref{sec:exp-zero-few}.

\section{Methodology}

\subsection{Problem Definition}
Let $D=\{(x_i=(d_i, t_i, w_i), y_i)\}_{i=1}^N$ denote $N$ examples, with input $x_i$ consisting of a document $d_i$, target $t_i$, and Wikipedia text $w_i$ about the target, and a stance label $y_i \in \{ \text{favor}, \text{against}, \text{neutral} \}$ as output. The goal is to infer  $y_i$ given $x_i$.

\subsection{Encoding Wikipedia Knowledge}
For the background knowledge, we use the raw text of Wikipedia pages instead of a Wikipedia knowledge graph because 1) a knowledge graph is more structured but inevitably suffers information loss when being constructed; \citet{liu2021enhancing} uses a commonsense knowledge graph to enhance stance detection, which is outperformed by our method that simply uses raw texts, as shown in Section \ref{sec:exp-zero-few}; 2) in addition, raw text is much more readily accessible and needs less preprocessing, especially for newly emerging targets.

To incorporate background knowledge about targets from Wikipedia, we propose \textbf{W}ikipedia \textbf{S}tance Detection BERT (WS-BERT). Depending on the textual style (formal vs. informal) of the documents, we introduce two variants of WS-BERT, namely WS-BERT-Single, for dealing with formal documents, and WS-BERT-Dual, for dealing with informal documents. Below we elaborate on the architectures of these models.

\begin{figure}[ht]
    \centering
    \includegraphics[width=0.4\textwidth]{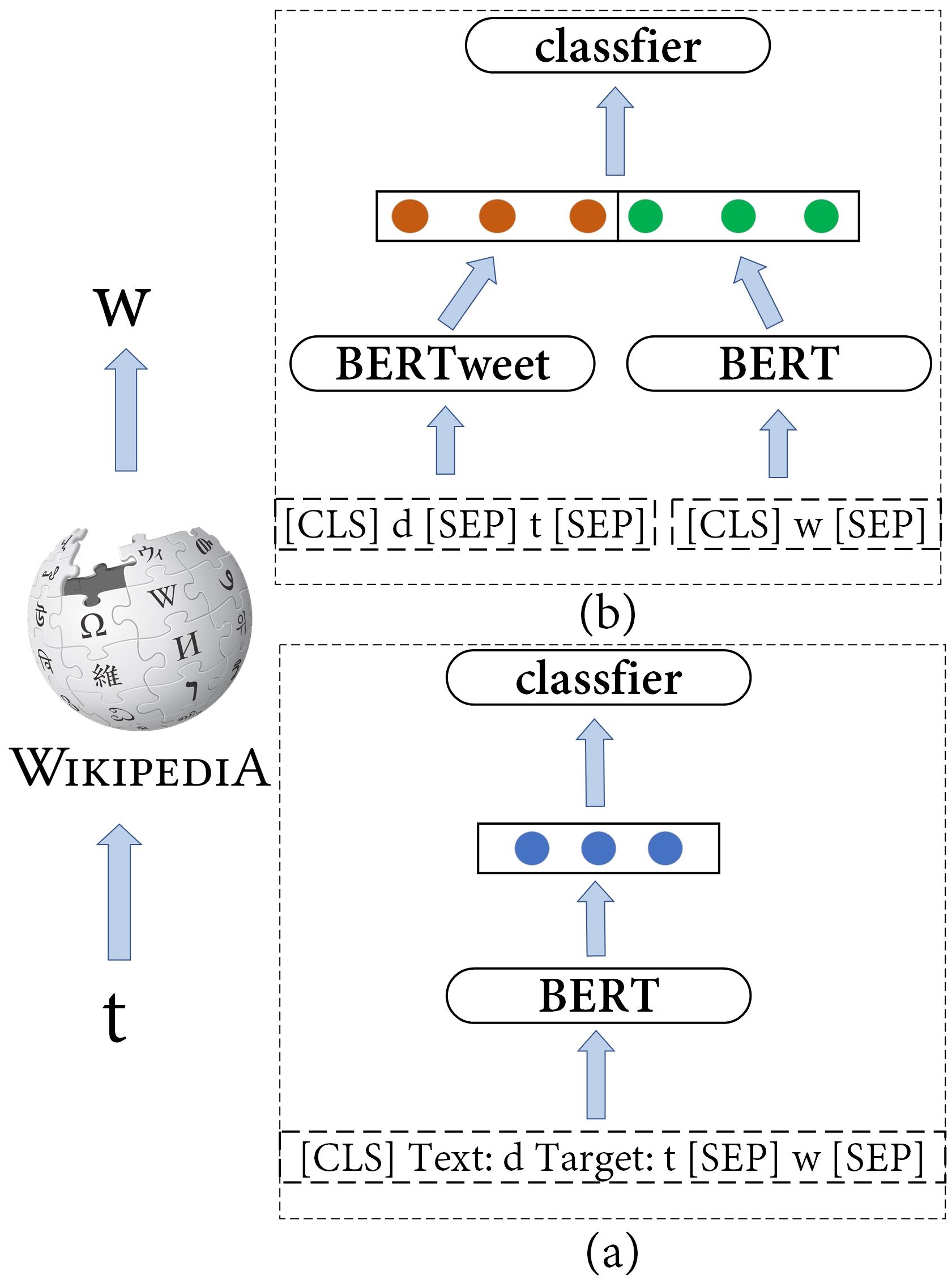}
    \caption{Architecture of (a) WS-BERT-Single and (b) WS-BERT-Dual.}
    \label{fig:ws-bert}
\end{figure}

\textbf{Infusing Wikipedia knowledge with formal documents.}
When documents are written in a formal style as Wikipedia articles, we use BERT that is also pretrained on Wikipedia articles to collectively encode the document $d$, the target $t$, and the Wikipedia knowledge $w$. Previous works \cite{allaway2021adversarial, liang2021target, li2021p, glandt2021stance, allaway2020zero, liu2021enhancing} treat the document $d$ and the target $t$ as a sequence pair and use BERT to encode it, with the input format as ``[CLS] $d$ [SEP] $t$ [SEP]''. Since BERT was originally designed to deal with at most two sequences\footnote{There do exist some works that have tried to make it encode three sequences simultaneously by using three [SEP] tokens \cite{xu2021fusing}.}, to encode the Wikipedia knowledge in addition to the document-target pair, we merge the document and the target into a single sequence and redesign the input format as ``[CLS] Text: $d$ Target: $t$ [SEP] $w$ [SEP]'' as shown in Figure \ref{fig:ws-bert}(a). 
Such an input format enables $d$, $t$, and $w$ to attend to each other during the encoding process. 
The pooled output of the final layer [CLS] embedding is used as the final representation of the input $x$.
Since one BERT is used, we call the model WS-BERT-Single.

\textbf{Infusing Wikipedia knowledge with informal documents.}
Social media has become a popular platform for people to express their views on public figures or political events. The opinions of online users are documented by noisy and casual user-generated texts. Such texts have a  different distribution than the Wikipedia corpus that BERT is pretrained on. In this case, we use BERTweet \cite{nguyen2020bertweet} or COVID-Twitter-BERT \cite{muller2020covid} that is pretrained on social media texts to encode the document-target pair, and use the vanilla BERT to encode Wikipedia knowledge, as shown in Figure \ref{fig:ws-bert}(b). We encode the document-target pair and the Wikipedia knowledge separately with two language models so as to minimize domain shift between the training examples used in this paper and the original pretraining corpora of the language models. 
We concatenate the two pooled outputs of the final layer [CLS] embeddings from two language models as the final representation of the input $x$.\footnote{Admittedly, concatenation of the two vectors seems naive, but it achieves satisfactory performance as shown in Section \ref{sec:target-specific} and \ref{sec:cross-target}; more sophisticated ways to fuse them like cross-attention count towards our future work.}
We call this model WS-BERT-Dual since we use two BERT-based language models.

\subsection{Stance Prediction}
The final representation from WS-BERT is fed into a single fully-connected layer and softmax layer to predict the stance label $\hat{y} \in \{ \text{favor}, \text{against}, \text{neutral} \}$, which is optimized by a cross-entropy loss.

\section{Experiments}

\subsection{Datasets}
\label{sec:datasets}
We evaluate the proposed WS-BERT model on three newly published datasets since 2020. For the targets in the three datasets, we use summaries of the fetched Wikipedia pages as the textual background knowledge.

P-Stance \cite{li2021p} is for target-specific and cross-target stance detection and it consists of tweets related to three politicians ``Biden'', ``Sanders'' and ``Trump''. We manually fetched the individual Wikipedia pages of the three politicians.

COVID-19-Stance \cite{glandt2021stance} is a dataset of pandemic-related tweets for target-specific stance detection and contains four targets: ``Anthony Fauci'', ``stay-at-home orders'', ``wear a face mask'', and ``keeping school closed''. The titles of the Wikipedia pages used are ``Anthony Fauci'', ``COVID-19 lockdowns'', ``Face masks during the COVID-19 pandemic in the United States'', and ``Impact of the COVID-19 pandemic on education''. Locating these Wikipedia pages is also a manual process.

Varied Stance Topics (VAST) \cite{allaway2020zero} is for zero/few-shot stance detection and comprises comments from \emph{The New York Times} ``Room for Debate'' section on a large range of topics covering broad themes. It has \textasciitilde6000 targets. We use an API\footnote{\url{https://pypi.org/project/wikipedia/}} to crawl the Wikipedia pages of them. For the targets that have multiple related Wikipedia pages, we choose the first one recommended by the API. For the targets that do not have any Wikipedia pages (\textasciitilde200, e.g., ``salt preference'' and ``tennis fans''), we use the targets themselves as background knowledge, with no additional information introduced.


\subsection{Evaluation Metric}
Following previous works \cite{mohammad2016semeval, mohammad2017stance}, we adopt macro-average of F1-score as the evaluation metric. For P-Stance where the examples only have two stance labels, $F_{\text{avg}} = (F_{\text{favor}} + F_{\text{against}}) /2 $. For COVID-19-Stance and VAST that have three stance labels, $F_{\text{avg}} = (F_{\text{favor}} + F_{\text{against}} +  F_{\text{neutral}} )/3 $.

\subsection{Experimental Setup}
We use WS-BERT-Dual in experiments on P-Stance and COVID-19-Stance, both of which consist of tweets. Following the setup in their original papers \cite{li2021p, glandt2021stance}, for P-Stance, we use BERTweet as the document-target encoder, and for COVID-19 Stance, we use COVID-Twitter-BERT as the the document-target encoder; for both datasets, BERT-base is used to encode Wikipedia knowledge. On VAST that comprises online debates, we use BERT-base to jointly encode the document-target-knowledge tuple.

All models are implemented using PyTorch.
The Wikipedia summaries are truncated to a maximum of 512 tokens.
We train the models using Adam optimizer with a batch size of $32$ for a maximum of $100$ epochs with patience of $10$ epochs. The weight decay is set to $5e-5$.
To speed up the training process we only finetune the top layers of the Wikipedia encoder in WS-BERT-Dual. We search the learning rate in $\{1e-5, 2e-5\}$ and the number of Wikipedia encoder layers to finetune in $\{1,2\}$.

On target-specific and zero/few-shot stance detection, we follow the standard train/validation/test splits of the three datasets. On cross-target stance detection, the model is trained on the train set of the source target, evaluated on the validation set of the source target, and tested on the combination of train, validation, and test set of the destination target, following the setup in P-Stance. The results are reported from the model with the best performance on the validation set.

\subsection{Target-specific Stance Detection}
\label{sec:target-specific}
For target-specific stance detection on P-Stance and COVID-19-Stance, we train a model separately for each target and test it on the same target.

\textbf{Baselines.} On P-Stance we compare to the baselines TAN \cite{du2017stance}, BiCE \cite{augenstein2016stance}, PGNN \cite{huang2018parameterized}, BERT, and BERTweet. On COVID-19-Stance we compare to TAN, ATGRU \cite{zhou2017connecting}, GCAE \cite{xue-li-2018-aspect}, COVID-Twitter-BERT, COVID-Twitter-BERT-NS \cite{xie2020self}, and COVID-Twitter-BERT-DAN \cite{xu2020dan}. 

\begin{table}[ht]
\begin{small}
\centering
\begin{tabular}{lcccc}
\hline
\textbf{Method} & \textbf{Trump} & \textbf{Biden} & \textbf{Sanders} & \textbf{Avg.} \\ \hline
TAN             & 77.1           & 77.6           & 71.6             & 75.1          \\
BiCE            & 77.2           & 77.7           & 71.2             & 75.4          \\
PGCNN           & 76.9           & 76.6           & 72.1             & 75.2          \\
GCAE            & 79.0           & 78.0           & 71.8             & 76.3          \\
BERT            & 78.3           & 78.7           & 72.5             & 76.5          \\
BERTweet        & 82.5           & 81.0           & 78.1             & 80.5          \\ \hline
BERTweet$\dagger$        & 85.2           & 82.5           & 78.5    & 82.1          \\
WS-BERT-Dual    & \textbf{85.8}  & \textbf{83.5}  & \textbf{79.0}             & \textbf{82.8} \\ \hline
\end{tabular}
\end{small}
\caption{Macro-average F1 scores of target-specific stance detection on P-Stance. BERTweet is implemented in \cite{li2021p} and BERTweet$\dagger$ is implemented in this paper.}
\label{tab:target-specific-pstance}
\end{table}


\begin{table}[ht]
\begin{small}
\addtolength{\tabcolsep}{-3pt}
\centering
\begin{tabular}{lccccc}
\hline
\textbf{Method}       & \textbf{Fauci} & \textbf{Home} & \textbf{Mask} & \textbf{School} & \textbf{Avg.}  \\ \hline
TAN                   & 54.7           & 53.6          & 54.6          & 53.4            & 54.1          \\
ATRGU                 & 61.2           & 52.1          & 59.9          & 52.7            & 56.5          \\
GCAE                  & 64.0           & 64.5          & 63.3          & 49.0            & 60.2          \\
CT-BERT           & 81.8           & 80.0          & 80.3          & 75.5            & 79.4          \\
CT-BERT-NS        & 82.1           & 78.4          & 83.3          & 75.3            & 79.8          \\
CT-BERT-DAN       & 83.2           & 78.7          & 82.5          & 71.7            & 79.0          \\ \hline
CT-BERT$\dagger$  & 83.0           & 83.6          & 83.8          & 81.7            & 83.0          \\
WS-BERT-Dual & \textbf{83.6}  & \textbf{85.0} & \textbf{86.6} & \textbf{82.2}   & \textbf{84.4} \\ \hline
\end{tabular}
\addtolength{\tabcolsep}{3pt}
\end{small}
\caption{Macro-average F1 scores of target-specific stance detection on COVID-19-Stance.
CT-BERT (short for \textbf{C}OVID-\textbf{T}witter-\textbf{BERT}) represents COVID-Twitter-BERT implemented in \cite{glandt2021stance} and CT-BERT$\dagger$ represents the model implemented in this paper.}
\label{tab:target-specific-covid-twi}
\end{table}

\textbf{Results and Analysis.} 
Results for P-Stance and COVID-19-Stance are shown in Table \ref{tab:target-specific-pstance} and Table \ref{tab:target-specific-covid-twi}. On P-Stance, BERTweet$\dagger$ outperforms the baselines on all targets, and WS-BERT-Dual further improves the performance and achieves the new state-of-the-art. On COVID-19-Stance, COVID-Twitter-BERT$\dagger$ outperforms all the baselines on targets except ``Fauci'', including the self-training baseline COVID-Twitter-BERT-NS and the domain adaptation baseline COVID-Twitter-BERT-DAN, both of which are trained using some additional external data. However, WS-BERT-Dual augmented with background knowledge outperforms state-of-the-art on all targets. 
Therefore, even on target-specific stance detection, where the models are fed sufficient data to learn the target, background knowledge about the target still helps improve performance.

\subsection{Cross-target Stance Detection}
\label{sec:cross-target}
We use P-Stance for cross-target stance detection, where the model is trained on one target, e.g., ``Trump',' and tested on another, e.g., ``Biden.''

\textbf{Baselines.}
We use BERTweet as a strong baseline, which is the most performant method reported in \cite{li2021p}.

\begin{table}[ht]
\begin{small}
\addtolength{\tabcolsep}{-3pt}
\centering
\begin{tabular}{lccc}
\hline
\textbf{Target}        & \multicolumn{1}{l}{BERTw} & \multicolumn{1}{l}{BERTw$\dagger$} & \multicolumn{1}{l}{WS-BERT-D} \\ \hline
\textbf{Trump$\rightarrow$Biden}   & 58.9  & 52.2  & \textbf{68.3} \\
\textbf{Trump$\rightarrow$Sanders} & 56.5  & 53.0  & \textbf{64.4} \\
\textbf{Biden$\rightarrow$Trump}   & 63.6  & 66.8  & \textbf{67.7} \\
\textbf{Biden$\rightarrow$Sanders} & 67.0  & 68.5  & \textbf{69.0} \\
\textbf{Sanders$\rightarrow$Trump} & 58.7  & 60.0  & \textbf{63.6} \\
\textbf{Sanders$\rightarrow$Biden} & 73.0  & 74.6  & \textbf{76.8} \\
\textbf{Avg.}          & 63.0  & 62.5  & \textbf{68.3} \\ \hline
\end{tabular}
\addtolength{\tabcolsep}{3pt}
\end{small}
\caption{Macro-average F1 scores of  cross-target stance detection on P-Stance. 
Trump$\rightarrow$Biden indicates that the model is trained on ``Donald Trump'' and tested on ``Joe Biden''.
BERTweet is implemented in \cite{li2021p} and BERTweet$\dagger$ is implemented in this paper.}
\label{tab:cross-target-pstance}
\end{table}

\textbf{Results and Analysis.} 
Results are shown in Table \ref{tab:cross-target-pstance}. We see that our implementation of BERTweet$\dagger$ outperforms BERTweet 
when the model is trained on ``Biden'' and ``Sanders''.
After infusing Wikipedia knowledge, WS-BERT-Dual enhances the performance on all six target pairs compared to BERTweet$\dagger$ and achieves the new state-of-the-art. 
Notably, the performance gains on ``Trump''$\rightarrow$``Biden'' and ``Trump''$\rightarrow$``Sanders'' are the biggest, which we argue is because the tweets about ``Trump'' mention the other two targets less, so that the model trained on ``Trump'' learns little knowledge transferable to the other two targets.  In this case, background knowledge about ``Biden'' or ``Sanders'' brings huge information gains, 
leading to substantial performance improvement.
In addition, compared to performance gain on target-specific stance detection, the gains in performance are more noticeable on this cross-target task, which signifies that background knowledge from Wikipedia is more important when the test target is outside of the training set.

\subsection{Zero-shot and Few-shot Stance Detection}
\label{sec:exp-zero-few}
Finally, we evaluate our model on zero-shot and few-shot stance detection using VAST, where the model is trained on thousands of targets and evaluated on targets that are not seen in the training data (zero-shot learning) and are seen just a few times in the training data (few-shot learning).

\textbf{Baselines.} 
We compare our model to BERT, TGA-Net \cite{allaway2020zero}, BERT-GCN \cite{lin2021bertgcn}, and CKE-Net \cite{liu2021enhancing}.

\begin{table}[ht]
\begin{small}
\centering
\begin{tabular}{lccc}
\hline
\multicolumn{1}{l}{\textbf{Method}} & \multicolumn{1}{l}{\textbf{Zero-shot}} & \multicolumn{1}{l}{\textbf{Few-shot}} & \multicolumn{1}{l}{\textbf{Overall}} \\ \hline
TGA-Net                            & 66.6                                   & 66.3                                  & 66.5                                 \\
BERT                                & 68.5                                   & 68.4                                  & 68.4                                 \\
BERT-GCN                            & 68.6                                   & 69.7                                  & 69.2                                 \\
CKE-Net                             & 70.2                                   & 70.1                                  & 70.1                                 \\ \hline
BERT$\dagger$                                & 70.1                                   & 70.0                                  & 70.0                                 \\
WS-BERT-Single                      & \textbf{75.3}                          & \textbf{73.6}                         & \textbf{74.5}                        \\ \hline
\end{tabular}
\end{small}
\caption{Macro-average F1 scores of zero-shot and few-shot stance detection on VAST. BERT is implemented in \cite{liu2021enhancing} and BERT$\dagger$ is implemented in this paper.}
\label{tab:zero-few-shot-vast}
\end{table}

\textbf{Results and Analysis.}
Results are shown in Table~\ref{tab:zero-few-shot-vast}. CKE-Net extracts the links between entities in targets and documents from a knowledge graph so as to make use of the commonsense knowledge. However, a well-finetuned BERT$\dagger$ implemented in this paper achieves performance on par with it, putting the effectiveness of CKE-Net into question. WS-BERT-Single significantly improves the performance on both zero-shot and few-shot learning by a huge margin, thus creating new state-of-the-art. We argue that such nontrivial performance gain is due to the presence of many targets in VAST that are difficult for the model to understand without background knowledge, such as ``b-12'' (a vitamin) and ``2big2fail''. 

As mentioned in Section \ref{sec:datasets}, the Wikipedia pages of the thousands of targets in VAST are retrieved by an API.
Admittedly, such an automated process might incur noisy information because the retrieved pages are not guaranteed to be the most relevant ones, and the summaries might miss useful content. However, even with the noise, our method manages to outperform the state-of-the-art baselines significantly, with an improvement in F1 of 4.5\%. Such a huge improvement demonstrates the robustness of our method in handling the noisy external knowledge: when the model is trained with noisy Wikipedia summaries, it learns to deal with such perturbations; as a result, during inference, with noisy external knowledge, it is still able to infer the correct stance. 

Moreover, the improvement on zero-shot learning is more observable compared to that on few-shot learning, because 
in few-shot learning the model is able to attend to some examples in the training data to understand the targets, while 
in zero-shot learning the model is not exposed to the targets at all, in which case background knowledge is of more importance.

\section{Conclusion}
In this paper we propose to utilize background knowledge about targets from Wikipedia to enhance stance detection. We propose WS-BERT with two variants to encode such knowledge. Such a simple yet effective method achieves state-of-the-art performance on three benchmark datasets and on three sub-tasks: in-target stance detection, cross-target stance detection, and zero/few-shot stance detection. The comprehensive and growing list of topics covered by Wikipedia ensures that our method will adapt to newly emerging targets. 

In the future, we plan to investigate incorporating knowledge about entities in the input documents, in addition to knowledge about the targets. Since Wikipedia pages may contain subjective opinions towards the targets, how to prevent the model from being negatively impacted by such bias when modeling the knowledge remains a promising research direction. Moreover, background knowledge from relevant news articles might also be helpful for inferring stances.

\section*{Acknowledgements}
We sincerely thank the reviewers for their insightful and constructive comments and suggestions that helped improve the paper.
This research was supported in part by DARPA under contract HR001121C0168.


\clearpage
\bibliography{anthology,custom}
\bibliographystyle{acl_natbib}

\clearpage
\appendix

\end{document}